\documentclass[letterpaper]{article} % DO NOT CHANGE THIS
\usepackage{aaai24}  % DO NOT CHANGE THIS
\usepackage{times}  % DO NOT CHANGE THIS
\usepackage{helvet}  % DO NOT CHANGE THIS
\usepackage{courier}  % DO NOT CHANGE THIS
\usepackage[hyphens]{url}  % DO NOT CHANGE THIS
\usepackage{graphicx} % DO NOT CHANGE THIS
\usepackage{natbib}  % DO NOT CHANGE THIS AND DO NOT ADD ANY OPTIONS TO IT
\usepackage{caption} % DO NOT CHANGE THIS AND DO NOT ADD ANY OPTIONS TO IT
\usepackage{multirow}
\usepackage{amsmath}
\usepackage{booktabs}
\usepackage[caption=false]{subfig}
\usepackage{algorithmic}
\usepackage[linesnumbered,ruled,vlined]{algorithm2e}
\SetKwInput{KwInput}{Input}
\SetKwInput{KwOutput}{Output}
\title{Differentially Private and Adversarially Robust Machine Learning:\\ An Empirical Evaluation}
\author{
    Janvi Thakkar\textsuperscript{\rm 1}, Giulio Zizzo\textsuperscript{\rm 2}, Sergio Maffeis \textsuperscript{\rm 1}\\
}
\affiliations{
    \textsuperscript{\rm 1}Department of Computing, Imperial College London\\
    \textsuperscript{\rm 2} IBM Research Europe
    janvi.thakkar22@imperial.ac.uk, giulio.zizzo2@ibm.com, sergio.maffeis@imperial.ac.uk
}

\usepackage{bibentry}
\begin{document}

\maketitle

\begin{abstract}
Malicious adversaries can attack machine learning models to infer sensitive information or damage the system by launching a series of evasion attacks. Although various work addresses privacy and security concerns, they focus on individual defenses, but in practice, models may undergo simultaneous attacks. This study explores the combination of adversarial training and differentially private training to defend against simultaneous attacks. While differentially-private adversarial training, as presented in DP-Adv~\cite{bu2021practical}, outperforms the other state-of-the-art methods in performance, it lacks formal privacy guarantees and empirical validation. Thus, in this work, we benchmark the performance of this technique using a membership inference attack and empirically show that the resulting approach is as private as non-robust private models. This work also highlights the need to explore privacy guarantees in dynamic training paradigms.
\end{abstract}

\section{Introduction} 
Despite their success, machine learning (ML) models are susceptible to various malicious attacks. 
Privacy attacks like model extraction \cite{tramer2016stealing} and membership inference attacks \cite{shokri2017membership} try to infer sensitive information of the model and its training datasets. 
Evasion attacks using adversarial samples \cite{szegedy2013intriguing} fool the model into predicting wrong outcomes, critically increasing the risks associated with ML models. 

%For instance, adversaries infer private information from the learned model parameters (privacy risks) or damage the system by launching a series of evasion attacks (security risks). This raises a significant concern when models are deployed in sensitive settings, and mitigating such risks becomes critical. 

Multiple studies have focused on devising techniques to preserve the privacy and security of ML models independently from each other. 
Several strategies were proposed \cite{chaudhuri2011differentially, kifer2012private, rubinstein2009learning} to defend against privacy attacks. One of the widely used approaches is to employ differential privacy in the training algorithm, DPSGD \cite{abadi2016deep}, to safeguard the privacy of individual datapoints. 

In response to evasion attacks, defense techniques such as adversarial training augment the training set with adversarial samples, generated using FGSM \cite{goodfellow2014explaining} or PGD \cite{madry2017towards}, significantly increasing the robustness of the resulting model.

There is little work addressing these concerns at the same time. 
%
% \needscitation\  observed that applying privacy or security measures developed independently can make a model more vulnerable to attacks. 
%
Preliminary research observed that applying DP makes the model more vulnerable to adversarial attacks \cite{tursynbek2020robustness}, and similarly, using adversarial training techniques makes the model more susceptible to privacy attacks \cite{shokri2017membership}.

Recent work \cite{phan2020scalable} proposed the StoBatch algorithm, which combines differential privacy and adversarial training to defend against simultaneous privacy and security attacks. The primary goal was to convert the training data to DP-private data and then leverage these DP-training samples to generate adversarial samples. However, this comes at the cost of model utility and robustness. 

DP-Adv~\cite{bu2021practical} tried addressing the limitations of StoBatch, by adopting the traditional DPSGD strategy \cite{abadi2016deep} and replacing each training sample with exactly one adversarial example. This approach is more practical, and provides improved utility and robustness compared to StoBatch.
% However, concerns were raised \needscitation\ over the privacy risk associated with the use of adversarial examples in the DP-Adv technique.
However, the approach has not been empirically evaluated. Furthermore there is a concern that owing to the use of adversarial samples, which are generated using a non-private optimizer, it may not be as private as DPSGD, i.e., a non-robust private model. 
% In this work, we investigate the effectiveness of the DP-Adv approach and provide experimental results to demonstrate its privacy.
% %The negative impact of privacy and robustness measures on models trained using either technique makes it challenging to combine the two approaches. 
%  \sm{Say briefly something about the method.} However, the performance degrades significantly in terms of model utility and robustness against evasion attacks. 
% %
% Subsequently, DP-Adv~\cite{bu2021practical} tries to address the limitations of StoBatch
% Although DP-Adv outperforms StoBatch, certain concerns have been raised over the privacy implications of generating the adversarial samples \needscitation. 
Thus, in this work, we address these privacy concerns, and empirically validate the claims made by the DP-Adv algorithm.
Our main contribution includes:
\begin{itemize}
    \item We benchmark the performance of the DP-Adv technique using a membership inference attack (MIA) and empirically showed its efficacy in preserving both the individual and group-level privacy of training data.
    \item We raise an important research question for the differential privacy community, which involves investigating privacy guarantees in the context of dynamic algorithms, i.e., constantly changing training paradigms.  
\end{itemize}

%\section{Related Work}

%Although several defenses were proposed to preserve the privacy and robustness of ML models, they concentrated on one particular domain. 

% We also try to address the questions raised regarding the privacy issues pertaining to the suggested technique. More details about the privacy risks and the approach is given in the later sections.
% \subsection{Our Contribution}
% The main contribution of our work includes benchmarking the performance of the DP-Adv approach on membership inference attacks and empirically showing its efficacy in preserving the privacy of training data. The performance observed is comparable to that of differentially private training of neural networks, ensuring both the individual and group-level privacy of training data. 

\section{Approach: DP-Adv}
In this section, we describe the DP-Adv algorithm \cite{bu2021practical}. The main intuition is to replace each normal training sample with exactly one adversarial example in the differentially private training, DPSGD algorithm. According to the post-processing property of DP, pre-processing the data has no impact on the privacy of the ML model; thus, the algorithm tackles the same optimization problem as proposed by \cite{madry2017towards} for adversarial training: \\
\begin{equation}
    \min_{\theta} \max_{\Delta : ||\Delta|| \leq \gamma} \mathcal{L}(f(x+\Delta; \theta), y)
\end{equation}
The algorithm uses non-DP optimizers to generate the adversarial samples, i.e., inner maximization problem; $x+\Delta$, where the $x$ is the normal training example and $\Delta$ is the optimal perturbation ($\gamma$ is the allowed perturbation), calculated using the PGD approach. For learning the $\theta$ parameters of the model (outer minimization problem), DP optimizers, such as DP-SGD or DP-Adam optimizer, are used.

The key steps of the approach are described below: 
\begin{enumerate} 
    \item \textbf{Pre-processing of training data:} For each epoch, we pre-process the normal training data and replace each data point in a batch with an adversarial example. In step 5, as described in appendix Algorithm~\ref{dp-adv-bu}, the attacker can be replaced by any adversarial training strategy, for instance, FGSM or PGD techniques.
    \item \textbf{DPSGD training:} After pre-processing the dataset, we apply the traditional DP-training, where we clip the per-sample gradient and apply the Gaussian mechanisms to add a scaled noise to the gradients.
    \item Above steps are iteratively performed until convergence.
\end{enumerate}
DP-Adv was reported to have higher utility and robustness performance against privacy and security attacks, surpassing the current SOTA represented by the StoBatch algorithm. While the research claims that replacing each training sample with exactly one adversarial sample does not affect the privacy of the method, one can argue that, we generate adversarial samples from benign samples using a non-private optimizer. Consequently, when we use all of these adversarial samples together to train our model, it may cause a group privacy problem, thereby leaking information about the training sample associated with the group. Similar concerns were also raised in \cite{OpenReview}. As a result, in this work, we use a membership inference attack to evaluate the individual and group-level privacy implications associated with the DP-Adv strategy.  
% In addition, there is this question with respect to the definition of differential privacy. For instance when we talk about the neighbouring datasets in DP, if any particular benign sample ($x$) is changed, it may affect all the adversarial samples generated using $x$. Thus, in this work we try to investigate the privacy risks associated with the proposed mechanisms by launching a membership inference attack. We provide detailed empirical evidence and analysis to address some of the above concerns. 
\section{Membership Inference Attack}
In this section, we provide an overview of the membership inference attack (MIA) and the corresponding technique we use to benchmark the performance of the DP-Adv method. The intuition behind the MIA is to infer whether any specific data point participated in the training set or not. It is observed in the literature that the vulnerability of the ML model towards MIA is directly proportional to the loss gap, which is the difference between the training and test loss \cite{sablayrolles2019white}. The loss gap also acts as an indicator to measure the degree of overfitting of the ML model. As we discussed earlier, adversarially trained models are more vulnerable to privacy attacks; \cite{rice2020overfitting} examined that this is due to the overfitting of robust models. In addition, \cite{yeom2018privacy} discovered that overfitting is sufficient for an attacker to launch a powerful MIA. 

The first formal design for carrying out the MIA was given by \cite{shokri2017membership}, where a classifier model was trained to differentiate between the training set members and non-members based on the output feature vector. This model is then utilized to launch a MIA on the target model. Another work \cite{yeom2018privacy} proposed a confidence-thresholding approach, where it classifies the membership of an input sample by comparing the confidence vector obtained from the target model with the pre-defined threshold value, usually learned using the shadow training technique \cite{shokri2017membership}. In this work, we use the thresholding technique similar to \cite{song2019privacy} to benchmark the performance of the DP-Adv approach. 
% Our goal is to investigate how the DP-Adv algorithm, which combines adversarial training and differential privacy (which independently shows vulnerability to each other), acts when it undergoes a membership inference attack.
%(\textbf{NOTE:} Section Membership Inference Performance and Utilizing Model's prediction on benign and adversarial examples need to be combined (currently shifted to appendix)).

\section{Experimental Setting}
\subsection{Dataset and Model}
We used MNIST \cite{deng2012mnist}, Fashion-MNIST \cite{xiao2017fashion}, and CIFAR10 \cite{krizhevsky2009learning} datasets to benchmark the performance of the four strategies, i.e., no-defense, only adversarial, only DP, and DP-Adv approach, when they undergo a MIA attack. For all three datasets, we use a 4-layer convolutional model comprising two consecutive convolutional blocks followed by two linear layers. The hyperparameters used for the experiments are defined in Table~\ref{tab:mia-hyperparameters} in the Appendix.

\section{Results and Analysis}
In this section, we benchmark the performance of the DP-Adv technique by evaluating the performance of membership inference attacks on individual and group-level data privacy. The term, \textit{individual level data privacy} refers to applying MIA on the entire dataset. In contrast, \textit{group level data privacy} pertains to evaluating inference attacks on specific groups or classes within the dataset. Please note, test accuracy measures the performance on the normal test dataset, and adversarial accuracy calculates the performance on the adversarially perturbed test dataset. 
%Please note, throughout the section, when we refer to training or test accuracy, we measure the micro accuracy, i.e., it checks the proportions of data points (across all the classes) that were classified correctly by the trained model.

\subsection{Performance}
% Please add the following required packages to your document preamble:
% \usepackage{multirow}
\begin{table}[h!]
\centering
\begin{tabular}{ c c cc cc }
\toprule
\multirow{2}{*}{\textbf{Dataset}} &
  \textbf{DP} &
  \multicolumn{2}{c }{\textbf{ADVTR}} &
  \multicolumn{2}{c }{\textbf{DP-ADVTR}} \\
 &
  \textbf{Test} &
  \multicolumn{1}{c }{\textbf{Test}} &
  \textbf{Adv} &
  \multicolumn{1}{c }{\textbf{Test}} &
  \textbf{Adv} \\ \midrule
MNIST   & 94.67 & \multicolumn{1}{|c|}{99.11} & 94.59 & \multicolumn{1}{| c|}{94.7}  & 73.15 \\
FMNIST  & 81.85 & \multicolumn{1}{|c|}{85.72} & 76.71 & \multicolumn{1}{| c|}{72.5}  & 46.99 \\
CIFAR10 & 54.62 & \multicolumn{1}{|c|}{65.97} & 33.69 & \multicolumn{1}{| c|}{43.08} & 24.37 \\ \bottomrule
\end{tabular}
\caption{Performance of the model with individual and simultaneous deployment of adversarial training and differential privacy technique. }
\label{tab:dp-adv-technique}
\end{table}
Table \ref{tab:dp-adv-technique} records the performance of the DP-Adv technique and its counterparts, i.e., only adversarial training and only DPSGD strategy. We can observe that for all the datasets, adversarial accuracy decreases for the DP-Adv technique compared to the adversarial training strategy. We attribute this to the integration of differential privacy, which involves gradient clipping and injecting noise at every iteration, thereby limiting the information learned and consequently decreasing the robustness. In the case of the FMNIST and CIFAR10 datasets, we observe a noticeable decrease in test accuracy for the combined interaction. However, the MNIST dataset\footnote{The analysis with respect to MNIST dataset is given in Appendix owing to the space constraints.} has no drop in test accuracy when we consider a minimum of two independent strategies. This is due to the complexity of the FMNIST and CIFAR10 dataset, where with the complex interaction of defense strategies, we observe a trade-off in the model's utility. 

\subsection{Individual Level Data Privacy}
In this section, we provide a detailed analysis of how models trained using different strategies, i.e., without any defense, using only differential privacy or adversarial training and the combined method DP-Adv, behave when they undergo MIA. We plot\footnote{Plots without smoothing can be found in Appendix.} the graphs for evaluating membership inference accuracy on original samples and additionally provide plots of training and test accuracy to observe how it influences the behavior of membership attacks.    
\subsubsection{On CIFAR10 dataset:}
\begin{figure}[h!]

\subfloat[Membership Inference Accuracy]{
  \includegraphics[width=\columnwidth]{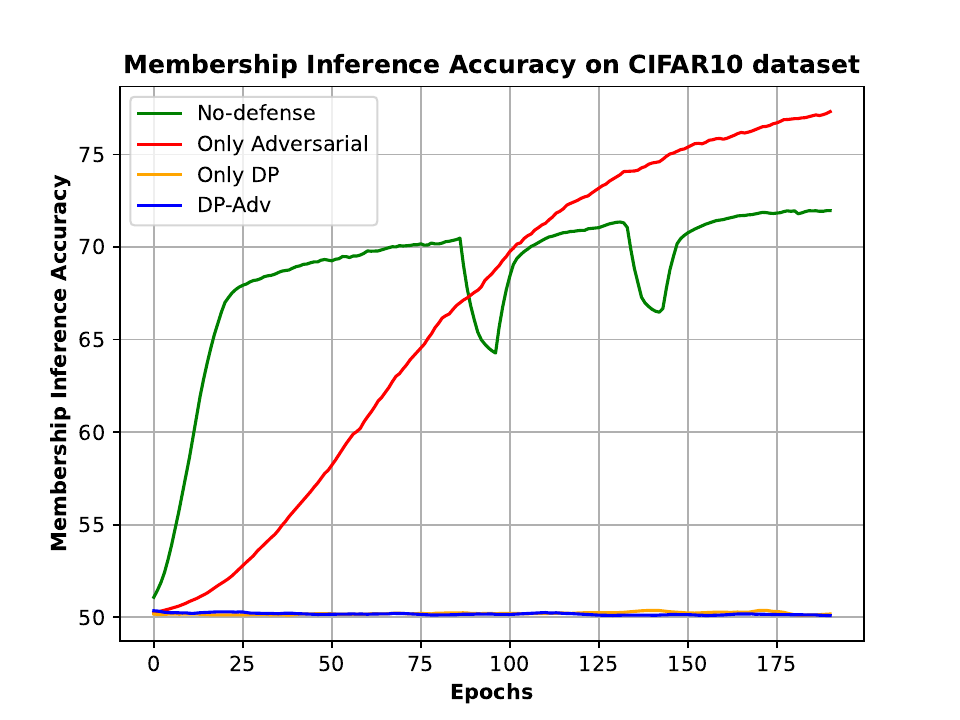}%
  \label{MIA_CIFAR10}
}

\subfloat[Comparison between Train and Test Accuracy]{
  \includegraphics[width=\columnwidth]{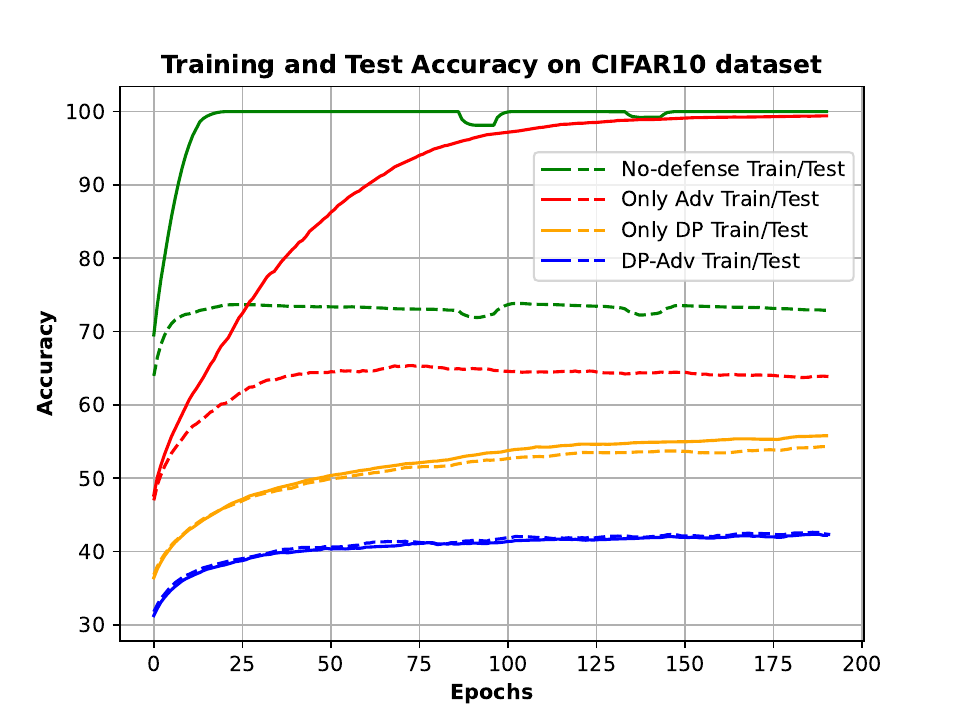}%
}%
\caption{Performance of the models trained using different techniques for 200 epochs on CIFAR10 dataset. The figures reports the value after applying running average smoothing with the window size of 10.}
\label{fig:mia-cifar10-plot}
\end{figure}
From Figure \ref{MIA_CIFAR10}, we can observe that the membership inference accuracy for the strategy trained with only adversarial training evidences a consistent monotonic increase; moreover, beyond 100 epochs, it surpasses the no-defense strategy. As discussed earlier, the vulnerability of robust training to privacy attacks is attributed to the difference between the training and test loss, commonly referred to as the loss gap. We report the results of the disparity between training and test accuracy (in Figure: \ref{fig:mia-cifar10-plot}b) and can observe that the training accuracy for only adversarial approach continues to increase while the test accuracy does not follow the same trajectory. Consequently, this widens the gap between the training and test accuracy and, thus, the increase in membership inference accuracy. The only adversarial training approach has a high susceptibility towards privacy attacks, however, when we apply the differentially private optimizer to the adversarial training (the DP-Adv approach), the membership inference accuracy drops. Specifically, we can see that for both the approach, only DP and DP-Adv, the training and test accuracy remains close, and the membership inference accuracy also remains consistently low. It is equivalent to having a random label assigned to each point, i.e., equal probability (50\%) of any sample being categorized as member or non-member. 
% Thus, based on this observation we can conclude that on individual level, the privacy of the training data is preserved, and that the DP-Adv is as private as non-robust private training.  
% \begin{figure}[h!]
%     \centering
%     \subfigure[\centering Membership Inference Accuracy]{{\includegraphics[width=8.1cm]{dp_adv_figures/Membership Inference Accuracy on MNIST dataset.pdf} }}%
%     % \hspace{0.05em}%
%     \subfigure[\centering Comparison between Train and Test Accuracy]{{\includegraphics[width=8.1cm]{dp_adv_figures/Training and Test Accuracy on MNIST dataset.pdf}}}%
%     \caption{Performance of the models trained using different techniques for 200 epochs on MNIST dataset. The figures reports the value after applying running average smoothing with the window size of 10.}%
%     \label{fig:mia-MNIST-plot}
% \end{figure}

\subsubsection{On Fashion-MNIST dataset:}
\begin{figure}[h!]

\subfloat[Membership Inference Accuracy]{
  \includegraphics[width=\columnwidth]{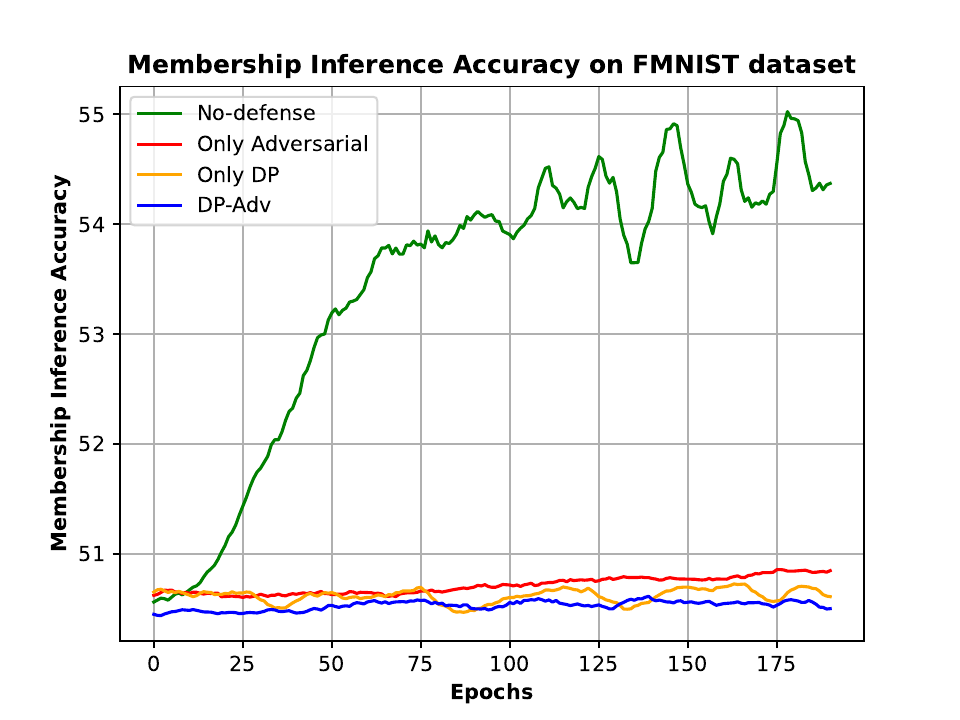}%
}

\subfloat[Comparison between Train and Test Accuracy]{
  \includegraphics[width=\columnwidth]{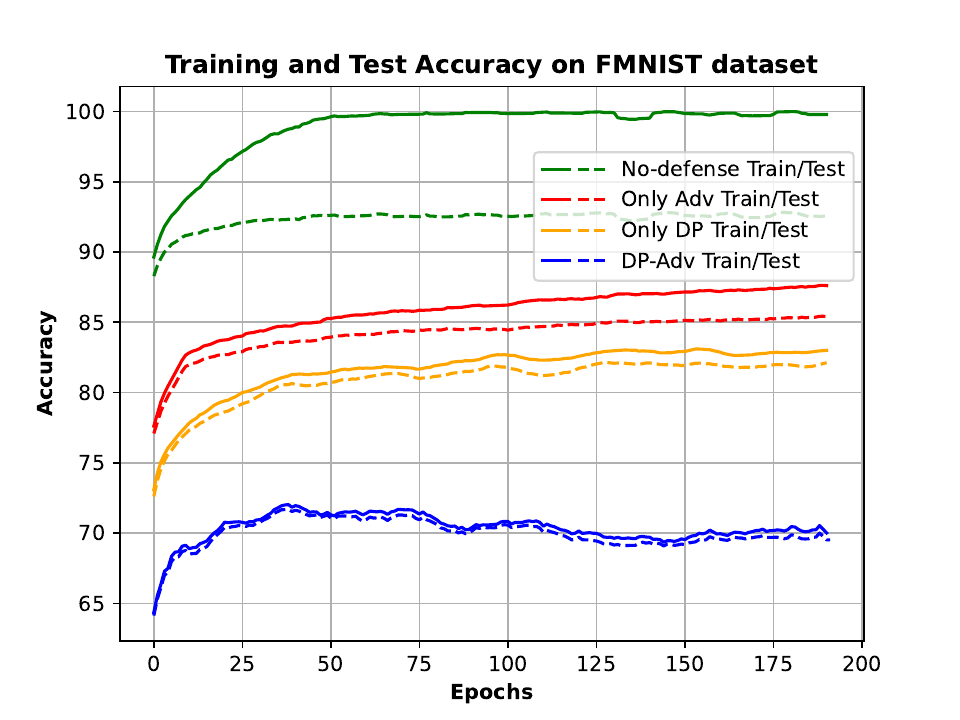}%
}%
\caption{Performance of the models trained using different techniques for 200 epochs on Fashion-MNIST dataset. The figures reports the value after applying running average smoothing with the window size of 10.}
\label{fig:mia-fmnist-plot}
\end{figure}
The Figure: \ref{fig:mia-fmnist-plot} represents the plots of the membership inference, training, and test accuracy for the model trained on various strategies for the Fashion-MNIST dataset. Unlike CIFAR10, we observe that for FMNIST, the membership inference accuracy for the adversarial approach is lower compared to the no-defense strategy. Simultaneously, the disparity between training and test accuracy for the no-defense strategy is high, affecting the membership inference accuracy. We know that robust models are more vulnerable to privacy attacks when they start to exhibit signs of overfitting. However, our analysis of training and test accuracy for adversarial training indicates that the model has not reached the point of overfitting. It's plausible that training for additional epochs may result in overfitting, altering the model's susceptibility to privacy attacks. Similar to the CIFAR10, the DP-Adv approach demonstrates low membership inference accuracy, implying its effectiveness against membership attacks. \\
\subsection{Group Level Data Privacy}
In the previous section, we observed the efficacy of the DP-Adv technique in preserving individual-level data privacy against the MIA. However, as mentioned earlier, group privacy issues may arise when all the different adversarial samples are used in the model training. The intuition is that when we use many adversarial samples generated from one particular training example, together they might result into leaking of more information about that specific point. To address this group privacy concern, we devised an experiment to see the behavior of membership inference attack when launched on the specific groups of the dataset. Since our model was trained using several adversarial examples associated with a benign example, consequently, we hypothesis individual classes as distinct groups and investigate whether they indeed expose more information about their training subgroup datasets. We compare the group-level membership inference accuracy between the non-robust private training (only DP) and the DP-Adv approach. We perform this comparison by inferring from both clean data groups and groups perturbed with adversarial techniques (more stronger attack owing to the use of information about the perturbation budget). 
%The strategy $\mathcal{S}_A$ launches more powerful attack as we use the knowledge about the perturbation budget. \\
\begin{figure}[h!]
\centering
\includegraphics[width = \linewidth]{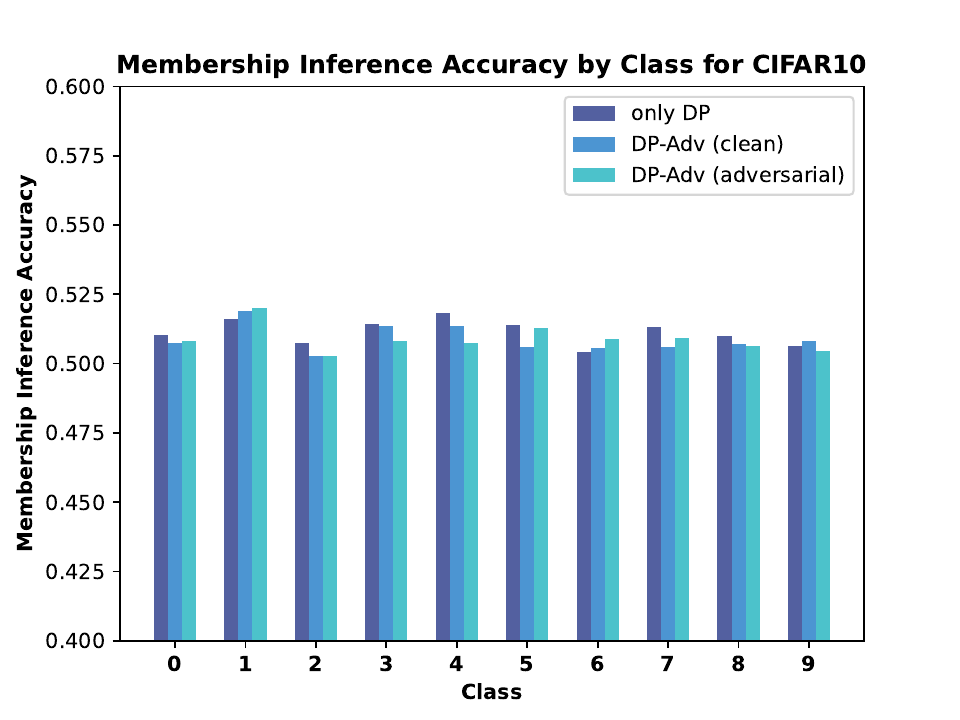}
\caption{Class wise membership inference accuracy on CIFAR10 dataset}
\label{fig:group-level-privacy}
\end{figure}

The Figure: \ref{fig:group-level-privacy} represents the performance of the MIA performed on the specific groups of the  CIFAR10 dataset. We can observe that overall, all three approach shows similar membership inference accuracy, within a consistent range of approximately 50\% to 52\%. Thus, despite the variations in the training methods and initial concerns regarding group privacy, the results indicate that the membership inference accuracy remains stable even with the adversarially perturbed groups. This further reinforces the protection of group privacy aspects of the DP-Adv approach. 

In summary, the empirical evidence suggests that the DP-Adv technique is as private as non-robust private models. While the concerns regarding the group level privacy gave the impression of additional privacy breach, this may not be the case. We attribute this to the dynamic changing of the dataset at every epoch. DPSGD ensured the privacy of individual data points by clipping the per-example gradient and adding the Gaussian noise to each update step in order to make them less observable. However, in the case of DP-Adv, at every iteration, we create a new dataset where each benign sample is replaced with the adversarial sample, i.e., the dataset changes dynamically at every epoch. Thus, it might lead to privacy properties that don't align perfectly with traditional differential privacy definitions. While we acknowledge that any deviation from established privacy definitions requires rigorous privacy analysis, our intuition serves as a starting point for a broader exploration of privacy guarantees in dynamic training paradigms.

\section{Conclusion}
To conclude, our empirical findings offer strong evidence suggesting that the DP-Adv approach offers privacy protection comparable to the models trained using only DP techniques. The experiments substantiate the safeguarding of individual and group-level data privacy. We acknowledge that while we provide the empirical evidence for DP-Adv approach, conclusive claims about its equivalent performance to the only DP strategy require rigorous privacy analysis. Finally, we highlight the need to explore privacy guarantees for constantly changing training algorithms.

\bibliography{aaai24}

\onecolumn
\section*{Appendix}\label{label:appendix}
\begin{algorithm}[]
\DontPrintSemicolon
  
  \KwInput{$\textbf{X} = \text{datapoints}\{x_1,x_2, ...., x_N\}$ \\
   % \textbf{Loss function}: $L(\theta) = \frac{1}{N} \sum_{i=1}^N L(\theta, x_i)$\\
  \textbf{Hyperparameters:} \\
  \textbf{Learning rate}: $\alpha$, \textbf{Noise scale:} $\sigma$, \\ \textbf{Clipping threshold} (gradient norm bound): $C$, \textbf{perturbation bound:} $\gamma$, \\
  \textbf{iterations:} $T$, \textbf{subsampling probability:} $q$, \textbf{initial weights:} $\theta_{0}$
}
% Select $k$ centroids $\textbf{S}^{(0)} = ({S}^{(0)}_{1},{S}^{(0)}_{2}, ..., {S}^{(0)}_{k})$ uniformly from X.
% \\
% $iterationForLloyd$ = number of iterations to run the algorithm.
% \\
 \For{$t = 0 \cdots T-1 $}{
    \text{Subsample a batch} $B_t \subseteq {1, \cdots, n}$ \text{with subsampling probability q}\\
    \For{each i $\in$ $B_t$}{
        \textbf{Generate Adversarial Example}\\
         $x_i \gets attacker (x_i, y_i, f; \gamma)$\\
         $g_i \gets \nabla_{\theta}\mathcal{L}(f(x_i, \theta_i), y_i)$\\
         \textbf{Clip the per-sample gradient}\\
         $g_i \gets g_i \cdot \min   \{1, \frac{C}{{||g_i||}_2}\}$\\
         % $\overline{g}_t(x_i) \gets \frac{g_t(x_i)}{\max \left(1, \frac{\lVert g_t(x_i) \rVert_2}{C}\right)}$\\
    }
    $g_t \gets \sum_{i \in B_t}g_i$ \\
    \textbf{Apply Gaussian Mechanisms}\\
   $g_t \gets g_t + \sigma C \cdot \mathcal{N}(0, \textbf{I})$ \\
   \textbf{Gradient descent step}\\
   $\theta_{t+1} \gets \theta_t - \frac{\alpha}{|B_t|} g_t$\\
 }
 \KwOutput{$\theta_T$ and compute the overall privacy cost $(\epsilon, \delta)$ using a privacy accounting method.}

\caption{Differentially Private Adversarial Training (DP-Adv) \cite{bu2021practical}}
\label{dp-adv-bu}
\end{algorithm}

\subsection{Hyperparameter}
% Please add the following required packages to your document preamble:
% \usepackage{multirow}
\begin{table}[h!]
\centering
\begin{tabular}{|c|c|ccc|}
\hline
\multirow{2}{*}{\textbf{Dataset}} & \textbf{DP Training} & \multicolumn{3}{c|}{\textbf{Adversarial Training}}                                                       \\
                                  & \textbf{dp\_epsilon} & \multicolumn{1}{c|}{\textbf{attack steps}} & \multicolumn{1}{c|}{\textbf{steps size}} & \textbf{epsilon} \\ \hline
\textbf{CIFAR10}       & 3 & \multicolumn{1}{c|}{10} & \multicolumn{1}{c|}{2/255} & 8/255 \\
\textbf{MNIST}         & 1 & \multicolumn{1}{c|}{25} & \multicolumn{1}{c|}{0.02}  & 0.25  \\
\textbf{Fashion-MNIST} & 1 & \multicolumn{1}{c|}{15} & \multicolumn{1}{c|}{0.02}  & 0.15  \\ \hline
\end{tabular}
\caption{Hyperparameters used for various datasets to evaluate the interplay between the differential privacy and adversarial training.}
\label{tab:mia-hyperparameters}
\end{table}
We used the hyperparameters defined in the \textbf{Table: \ref{tab:mia-hyperparameters}} for training our model separately using DP training and Adversarial training. When adopting the DP-Adv approach, we combine the parameter settings from both individual methods. Furthermore, all the models were trained for a total of $200$ epochs, with a learning rate of $0.005$ and weight decay of $5e-4$. \\
\begin{figure}[h!]

\subfloat[Membership Inference Accuracy]{
  \includegraphics[width=0.5\columnwidth]{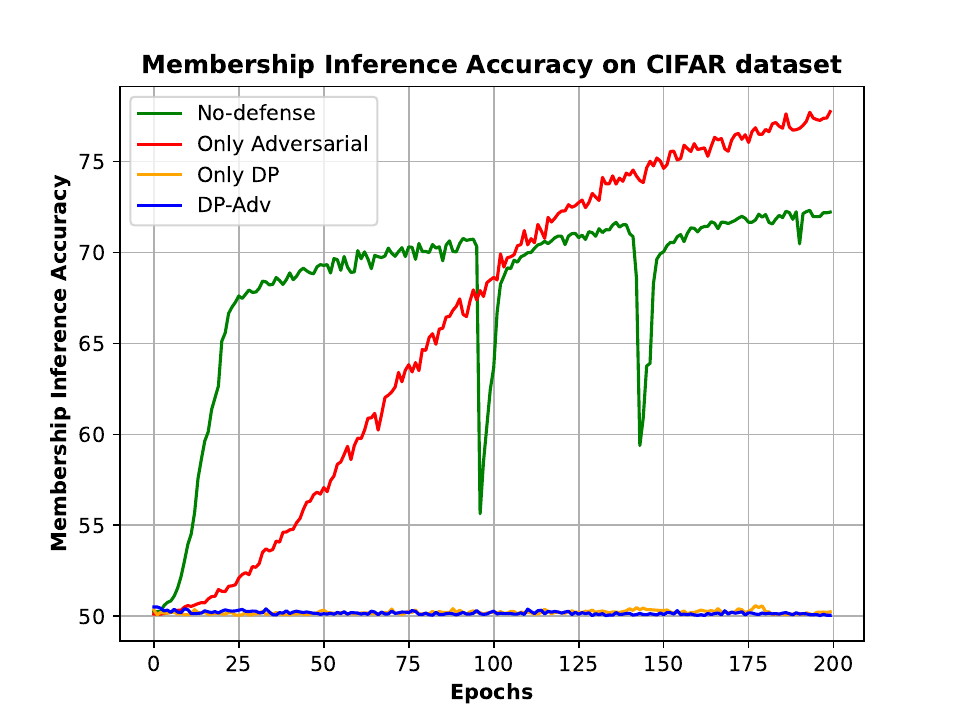}%
}
\subfloat[Comparison between Train and Test Accuracy]{
  \includegraphics[width=0.5\columnwidth]{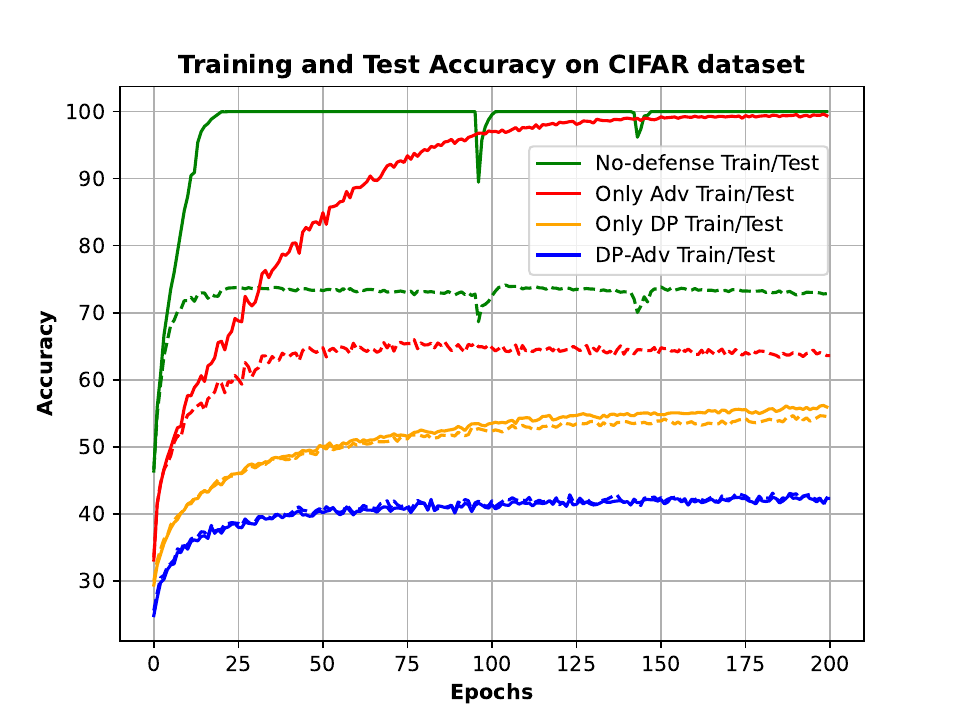}%
}%
\caption{Performance of the models trained using different techniques for 200 epochs on CIFAR10 dataset}
\label{fig:mia-CIFAR10-plot-ns}
\end{figure}

\subsection{CIFAR10: MIA Table}
% Please add the following required packages to your document preamble:
% \usepackage{multirow}
\begin{table}[h!]
\centering
\begin{tabular}{|c|cccc|cccc|}
\hline
\multirow{2}{*}{\textbf{Technique}} &
  \multicolumn{4}{c|}{\textbf{MI performance on Benign Examples}} \\
 &
  \multicolumn{1}{c|}{\textbf{Accuracy}} &
  \multicolumn{1}{c|}{\textbf{Precision}} &
  \multicolumn{1}{c|}{\textbf{Recall}} &
  \textbf{F1-Score} \\ \hline
no-defense &
  \multicolumn{1}{c|}{72.19} &
  \multicolumn{1}{c|}{65.05} &
  \multicolumn{1}{c|}{95.91} &
  77.52 \\
only-adv &
  \multicolumn{1}{c|}{77.74} &
  \multicolumn{1}{c|}{75.26} &
  \multicolumn{1}{c|}{82.66} &
  78.79 \\
only-dp &
  \multicolumn{1}{c|}{50.23} &
  \multicolumn{1}{c|}{50.19} &
  \multicolumn{1}{c|}{62.51} &
  55.67 \\
DP-Adv &
  \multicolumn{1}{c|}{50.08} &
  \multicolumn{1}{c|}{51.91} &
  \multicolumn{1}{c|}{99.10} &
  68.13 \\ \hline
\end{tabular}
\caption{Membership Inference (MI) Performance on CIFAR10 dataset.}
\label{tab:mia-cifar10}
\end{table}
The \textbf{Table: \ref{tab:mia-cifar10}} reports the results of membership inference accuracy, precision, recall, and F1-score with respect to the positive class. In the context of a MIA, having both a high F1-score and membership inference accuracy implies that the attacker is able to identify the correct members of the training set (high recall) and is making accurate predictions (high precision), thereby suggesting that it becomes easier for an adversary to infer the information, compromising the privacy of the model. Similarly, the low F1-score represents that the adversary is not easily able to distinguish between the presence or absence of data points, thus providing more privacy to the model. From the \textbf{Table: \ref{tab:mia-cifar10}}, we can observe that for no-defense and only adversarial strategy, the F1 score is high compared to other techniques, implying that the membership can be effectively distinguished for these models. For only DP and DP-Adv strategy, the overall F1-score is low; although we observe high recall (for DP-Adv), the precision and membership inference accuracy are quite low and thus make it challenging for the adversaries to infer the membership.

\subsection{Fashion-MNIST: MIA Table}
% Please add the following required packages to your document preamble:
% \usepackage{multirow}
% \usepackage[table,xcdraw]{xcolor}
% If you use beamer only pass "xcolor=table" option, i.e. \documentclass[xcolor=table]{beamer}
\begin{table}[h!]
\centering
\begin{tabular}{|c|cccc|cccc|}
\hline
 &
  \multicolumn{4}{c|}{\textbf{MI Performance on Benign Examples}} \\
\multirow{-2}{*}{\textbf{Technique}} &
  \multicolumn{1}{c|}{\textbf{Accuracy}} &
  \multicolumn{1}{c|}{\textbf{Precision}} &
  \multicolumn{1}{c|}{\textbf{Recall}} &
  \textbf{F1-Score} \\ \hline
no-defense &
  \multicolumn{1}{c|}{54.87} &
  \multicolumn{1}{c|}{52.63} &
  \multicolumn{1}{c|}{96.66} &
  68.15 \\
only-adv &
  \multicolumn{1}{c|}{50.87} &
  \multicolumn{1}{c|}{50.66} &
  \multicolumn{1}{c|}{66.56} &
  57.53 \\
only-dp &
  \multicolumn{1}{c|}{50.58} &
  \multicolumn{1}{c|}{50.54} &
  \multicolumn{1}{c|}{61.43} &
  55.46 \\
DP-Adv &
  \multicolumn{1}{c|}{50.53} &
  \multicolumn{1}{c|}{50.31} &
  \multicolumn{1}{c|}{72.89} &
  59.53 \\ \hline
\end{tabular}
\caption{Membership Inference Performance on Fashion-MNIST dataset.}
\label{tab:mia-fmnist}
\end{table}
The \textbf{Table: \ref{tab:mia-fmnist}} reports the membership inference accuracy, precision, recall, and F1-score for the positive class across different strategies tested on the Fashion-MNIST dataset. Overall, the membership inference accuracy and F1-score are low for all the strategies except for the no-defense strategy, which is slightly more vulnerable to privacy attacks. DP-Adv also shows comparable results to that of only DP strategy, indicating that the model trained using DP-Adv achieves a similar privacy level as that of non-robust private training. \\
\begin{figure}[h!]

\subfloat[Membership Inference Accuracy]{
  \includegraphics[width=0.5\columnwidth]{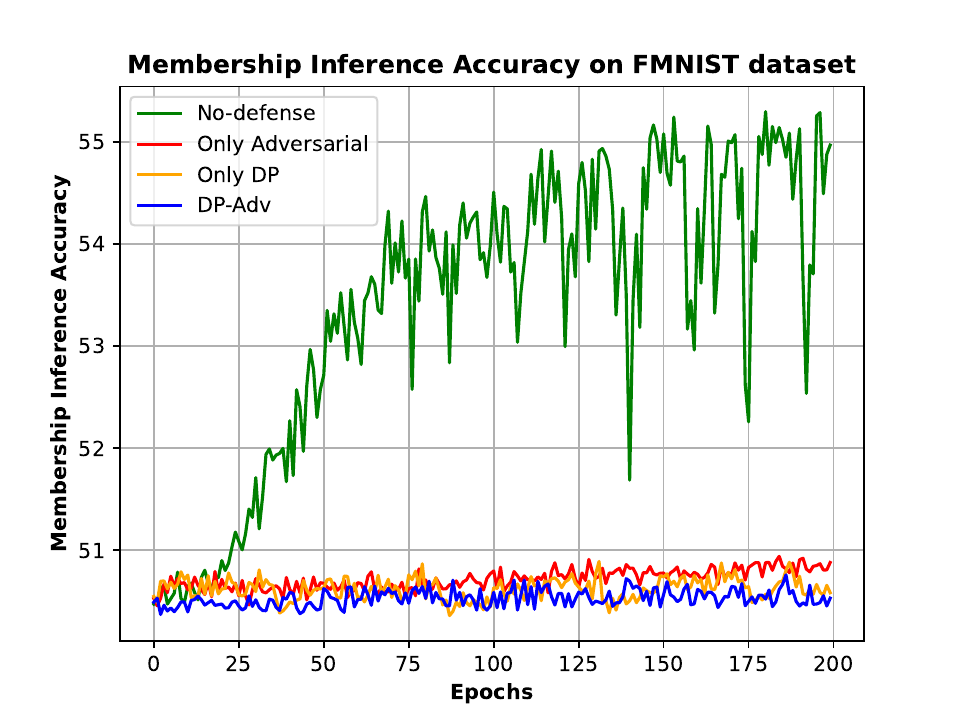}%
}
\subfloat[Comparison between Train and Test Accuracy]{
  \includegraphics[width=0.5\columnwidth]{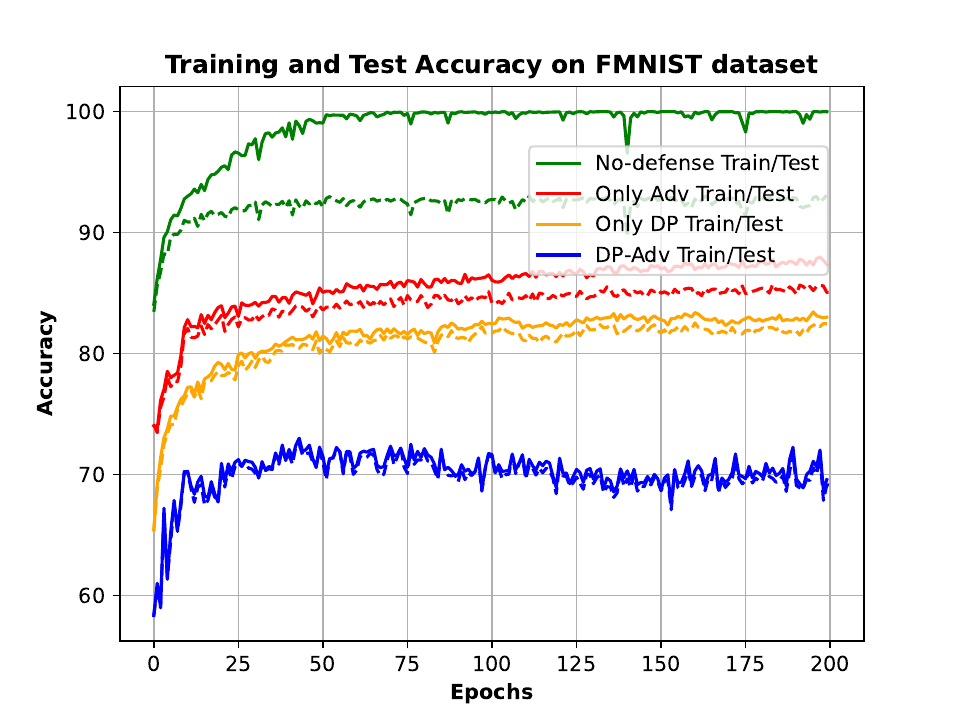}%
}%
\caption{Performance of the models trained using different techniques for 200 epochs on FMNIST dataset}
\label{fig:mia-fmnist-plot-ns}
\end{figure}

\subsection{MNIST: MIA Table}
% Please add the following required packages to your document preamble:
% \usepackage{multirow}
% \usepackage[table,xcdraw]{xcolor}
% If you use beamer only pass "xcolor=table" option, i.e. \documentclass[xcolor=table]{beamer}
\begin{table}[h!]
\centering
\begin{tabular}{|c|cccc|cccc|}
\hline
 &
  \multicolumn{4}{c|}{\textbf{MI Performance on Benign Examples}}  \\
\multirow{-2}{*}{\textbf{Technique}} &
  \multicolumn{1}{c|}{\textbf{Accuracy}} &
  \multicolumn{1}{c|}{\textbf{Precision}} &
  \multicolumn{1}{c|}{\textbf{Recall}} &
  \textbf{F1-Score} \\ \hline
no-defense &
  \multicolumn{1}{c|}{50.80} &
  \multicolumn{1}{c|}{50.41} &
  \multicolumn{1}{c|}{98.94} &
  66.79 \\
only-adv &
  \multicolumn{1}{c|}{51.73} &
  \multicolumn{1}{c|}{50.92} &
  \multicolumn{1}{c|}{95.50} &
  66.43 \\
only-dp &
  \multicolumn{1}{c|}{50.01} &
  \multicolumn{1}{c|}{50.00} &
  \multicolumn{1}{c|}{99.93} &
  66.65 \\
DP-Adv &
  \multicolumn{1}{c|}{50.37} &
  \multicolumn{1}{c|}{50.53} &
  \multicolumn{1}{c|}{34.90} &
  41.29 \\ \hline
\end{tabular}
\caption{Membership Inference Performance on MNIST dataset.}
\label{tab:mia-mnist}
\end{table}
The \textbf{Table: \ref{tab:mia-mnist}} reports the membership inference accuracy, precision, recall, and F1-Score of the positive class for the different strategies tested on the MNIST dataset. Overall, the membership inference accuracy and precision are low for all the strategies. Although the membership inference accuracy is slightly higher in the case of DP-Adv when compared to only the DP approach, its recall is extremely low, implying that the attack is potentially misclassifying the training points as non-members. Furthermore, the F1-score of DP-Adv is also low, indicating its effectiveness against the membership inference attack. Overall, based on this observation, we can say that on an individual level, the privacy of the training data is preserved and that the DP-Adv is as private as non-robust private training.

\subsubsection{Performance on MNIST dataset}
\begin{figure}[h!]

\subfloat[Membership Inference Accuracy]{
  \includegraphics[width=0.5\columnwidth]{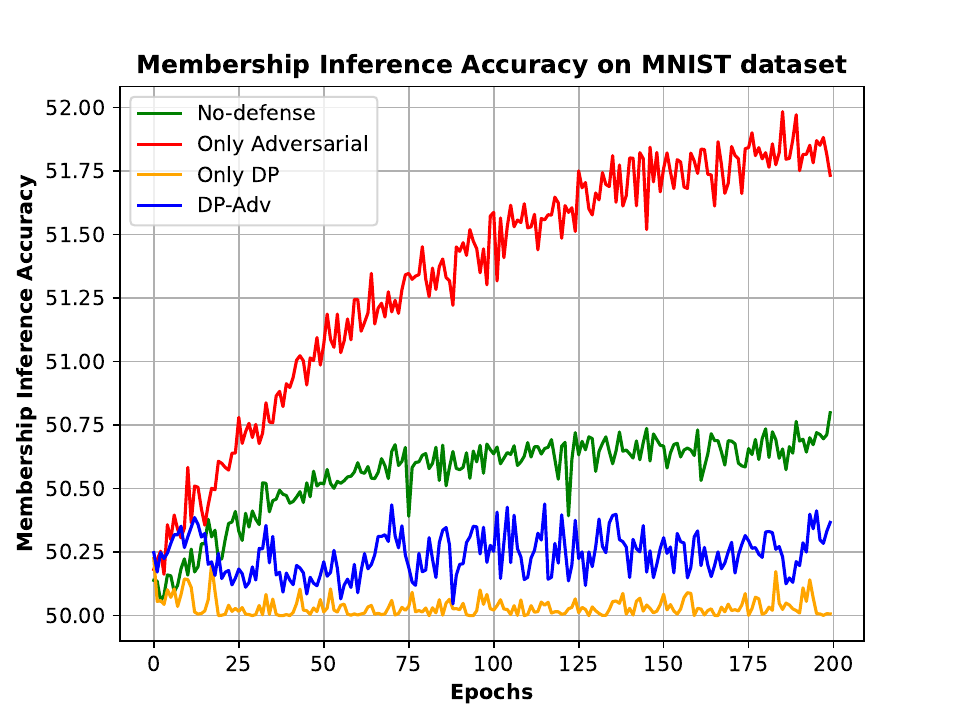}%
}
\subfloat[Comparison between Train and Test Accuracy]{
  \includegraphics[width=0.5\columnwidth]{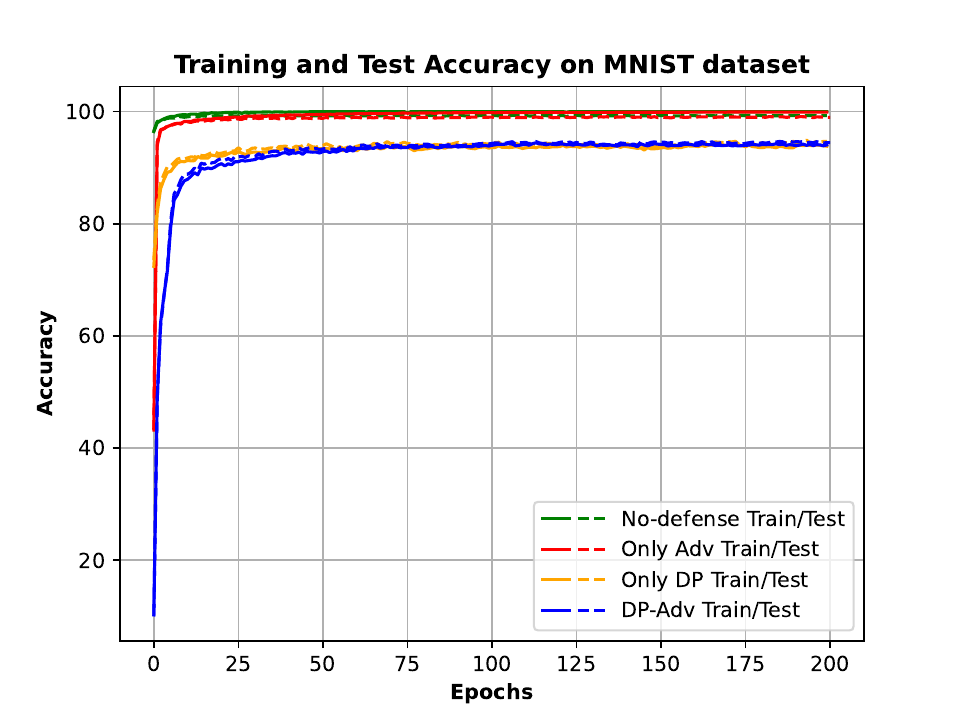}%
}%
\caption{Performance of the models trained using different techniques for 200 epochs on MNIST dataset.}
\label{fig:mia-mnist-plot}
\end{figure}
For the MNIST dataset (\textbf{Figure: \ref{fig:mia-mnist-plot}}, we can observe an increase in membership inference accuracy for adversarial training compared to other techniques. However, quantitatively, the values are still around 52\%, representing only a minor improvement over the baseline strategy (random guessing). We believe this to be because of the simplicity of the MNIST dataset, which, even in cases of potential overfitting, retains the ability to generalize well due to its straightforward nature. Thus, models trained using the MNIST dataset generalize well, resulting in small disparities between training and test accuracy (\textbf{Figure: \ref{fig:mia-mnist-plot}}). The results are in contrast to the CIFAR10 dataset, which is complex and potentially leads to overfitting of the model and larger disparities. Consequently, it may not generalize well on the unseen dataset, making them more vulnerable to inference attacks. \\

\end{document}